\documentclass[letterpaper]{article} % DO NOT CHANGE THIS
\usepackage{aaai2026}  % DO NOT CHANGE THIS
\usepackage{times}  % DO NOT CHANGE THIS
\usepackage{helvet}  % DO NOT CHANGE THIS
\usepackage{courier}  % DO NOT CHANGE THIS
\usepackage[hyphens]{url}  % DO NOT CHANGE THIS
\usepackage{graphicx} % DO NOT CHANGE THIS
\usepackage{natbib}  % DO NOT CHANGE THIS AND DO NOT ADD ANY OPTIONS TO IT
\usepackage{caption} % DO NOT CHANGE THIS AND DO NOT ADD ANY OPTIONS TO IT
\usepackage{algorithm}
\usepackage{algorithmic}

\usepackage{xcolor}
\usepackage{soul}
\soulregister{\cite}7

\usepackage{graphicx}

\usepackage{booktabs}
\usepackage{multirow}
\usepackage{adjustbox}
\usepackage[skins]{tcolorbox}

\usepackage{graphicx}
\usepackage{xcolor}

\newcommand{\answerYes}[1]{\textcolor{blue}{#1}} 
\newcommand{\answerNo}[1]{\textcolor{teal}{#1}} 
\newcommand{\answerNA}[1]{\textcolor{gray}{#1}}

\newcommand{\jy}[1]{{\color{magenta} [\bf {JY:}  #1]}}

\newcommand{\rparagraph}[1]{\vspace{1.2mm}\noindent\textbf{#1}}
\tcbset{highlight style/.style={colback=yellow!30, sharp corners, boxrule=0pt, borderline west={1pt}{0pt}{yellow!80!black}}}

\usepackage{newfloat}
\usepackage{listings}
\DeclareCaptionStyle{ruled}{labelfont=normalfont,labelsep=colon,strut=off} % DO NOT CHANGE THIS
\floatstyle{ruled}
\newfloat{listing}{tb}{lst}{}
\floatname{listing}{Listing}
\newtcolorbox{mybox}{
  enhanced,
  colback=lightgray!15, % Changed the background color to light gray
  colframe=lightgray!15!, % Changed the frame color to light gray
  fonttitle=\bfseries,
  colbacktitle=lightgray!85!black, % Changed the title background color to light gray
  coltitle=white,
  attach boxed title to top left={yshift=-2mm, xshift=3mm},
  boxed title style={sharp corners},
  arc=1mm,
  drop fuzzy shadow
}

\title{Collaborative Evaluation of Deepfake Text with Deliberation-Enhancing Dialogue Systems}
\author{
    Jooyoung Lee\textsuperscript{\rm 1}, 
    Xiaochen Zhu\textsuperscript{\rm 2}, 
    Georgi Karadzhov\textsuperscript{\rm 2} \\
    Tom Stafford\textsuperscript{\rm 3},
    Andreas Vlachos\textsuperscript{\rm 2},
    Dongwon Lee\textsuperscript{\rm 1}
}
\affiliations{
    \textsuperscript{\rm 1}The Pennsylvania State University, USA, 
        \textnormal{\{jfl5838, dongwon\}@psu.edu}  \\
    \textsuperscript{\rm 2} University of Cambridge, United Kingdom, 
        \textnormal{\{xz479, gmk34, av308\}@cam.ac.uk}  \\
    \textsuperscript{\rm 3} University of Sheffield, United Kingdom, \textnormal{t.stafford@sheffield.ac.uk}
   
}

\usepackage{bibentry}
\begin{document}

\maketitle

\begin{abstract}
The proliferation of generative models has presented significant challenges in distinguishing authentic human-authored content from deepfake content. Collaborative human efforts, augmented by AI tools, present a promising solution. In this study, we explore the potential of DeepFakeDeLiBot, a deliberation-enhancing chatbot, to support groups in detecting deepfake text. Our findings reveal that group-based problem-solving significantly improves the accuracy of identifying machine-generated paragraphs compared to individual efforts. While engagement with DeepFakeDeLiBot does not yield substantial performance gains overall, it enhances group dynamics by fostering greater participant engagement, consensus building, and the frequency and diversity of reasoning-based utterances. Additionally, participants with higher perceived effectiveness of group collaboration exhibited performance benefits from DeepFakeDeLiBot. These findings underscore the potential of deliberative chatbots in fostering interactive and productive group dynamics while ensuring accuracy in collaborative deepfake text detection. The  experiment datasets are available at: https://github.com/Brit7777/icwsm26-deepfakedelibot.
\end{abstract}

\section{Introduction}

%\begin{figure}[h!]
%    \centering
%     \includegraphics[width=0.47\textwidth]{figures/rqs_new.png}
%    \caption{Proposed research questions.}
%    \label{fig:rqs}
%\end{figure}

%With the advances in Artificial intelligence (AI) technologies, 
Large language models (LLMs) have transformed people's writing practice with remarkably fluent and human-like generation capabilities. 
However, their use in real-world applications is hampered by issues such as preserving biases or stereotypes \cite{kotek2023gender}, spreading mis/disinformation \cite{lucas-etal-2023-fighting, barman2024dark}, and facilitating plagiarism \cite{lee2023language, hutson2024rethinking}. To this end, researchers have recently been putting efforts into distinguishing deepfake texts (i.e., texts generated by machines) from human-authored texts. In particular, the focus has been on the curation of detection benchmarks \cite{uchendu-etal-2021-turingbench-benchmark, li2024mage, wang-etal-2024-m4} and the automation of detection procedure \cite{venkatraman-etal-2024-gpt, hu2023radar, wang-etal-2023-seqxgpt, mitchell2023detectgpt}. Yet, these detectors can be easily fooled by simple paraphrasing \cite{krishna2024paraphrasing} and are not robust to unseen models and domains \cite{weber2023testing}. These limitations necessitate exploring alternative strategies, such as integrating human-in-the-loop mechanisms, where human evaluators validate or supplement existing detectors.

Prior studies \cite{uchendu-etal-2021-turingbench-benchmark, dou-etal-2022-gpt, jakesch2023human} have reported that humans alone struggle to differentiate deepfake texts, performing only slightly better than random guessing. Even with additional training with examples and instructions, the performance gain was limited \cite{clark-etal-2021-thats}. While much of the existing literature has focused on individual-based detection, the role of collaborative problem-solving in tackling deepfake text remains underexplored. This gap is particularly consequential given that  detecting deepfake content is a cognitively demanding task that extends beyond surface-level recognition. It requires individuals to engage in sophisticated reasoning and critically evaluate ambiguous or potentially deceptive information.\citet{uchendu2023does} provide a first step by demonstrating that group collaboration can improve detection accuracy by 10 to 15\% over individual performance. This naturally opens up a new question: \textit{what makes group deliberation effective in such cognitively demanding, adversarial settings—and how might we further support it, particularly with the help of AI systems?}

%While the majority of literature focused on investigating individual-based deepfake detection, the role of collaborative problem-solving in deepfake detection is understudied. To the best of our knowledge, \citet{uchendu2023does} is the first to examine how groups perform on this task. Consistent with prior research in collective intelligence, they found that group collaboration enhances participants' detection of deepfakes by 10 to 15 percent compared to the performance of individual members. This naturally opens up a new question: \textit{what makes group deliberation successful and how can we further facilitate the collaborative deepfake detection?}

%A long history of empirical work has shown that discussion results may be influenced by several factors. 
Facilitating constructive and balanced discussion can be challenging due to several factors. For example, not all people are willing to actively participate in discussions. Additionally, some people may solely seek information consistent with their own perspectives, which can make it difficult for them to understand or respect others’ contrasting viewpoints \cite{stromer2009agreement}. Human moderators can play an important role in bridging the aforementioned gaps. Yet, due to the synchronous nature of online chat, moderators face a high managerial overhead in tasks like discussion stage management, opinion summarization, and consensus-building support. The research community has attempted to develop systems that can assist human moderation \cite{lee2020solutionchat} or an artificial moderator that can completely replace humans' involvement \cite{kim2021moderator, karadzhov2024delibots}. Not limited to moderation, a system to support reasoned argumentation \cite{drapeau2016microtalk} and a consensus-building \cite{shin2022chatbots} have also been explored. 

Building on this line of work, we integrate a deliberation-enhancing dialogue agent into group discussions and investigate its role in the domain of deepfake text detection—a high-stakes task where reasoning and collaboration are critical. Rather than moderating or classifying content, our bot prompts reflection, encourages balanced participation, and supports group reasoning dynamics. Through this setup, we aim to address three research questions (RQs): (1) \textbf{RQ1}: How does a deliberation-enhancing bot affect the individuals' performance of deepfake text detection?; (2) \textbf{RQ2}: How does the involvement of a deliberation-enhancing bot affect collaboration dynamics (e.g., engagement, even participation, consensus formation, probing dynamics and change of minds)?; (3) \textbf{RQ3}: What conditions make group collaboration with a deliberation-enhancing bot effective?.
%\lee{DeepFakeDeLiBot is mentioned without proper introduction}

In this work, we first curate a set of 14 articles, each consisting of three paragraphs: two written by humans and one generated by GPT-2 \cite{radford2019language} or GPT-3.5. We also introduce DeepFakeDeLiBot, a deliberation-enhancing bot for deepfake text detection, which is built upon the dialogue system presented by \citet{karadzhov2024delibots}. DeepFakeDeLiBot \cite{karadzhov2024delibots} prompts users with questions that foster collaborative discussion \textit{without} providing task-specific solutions or knowledge. This design allows us to isolate and analyze the effects of the deliberation process itself, free from interference by the bot offering correct answers.
% To customize DeepFakeDeLiBot for seamless deepfake text detection, we developed our dialogue system, DeepFakeDeLiBot, by expanding the candidate utterance pools for the bot to retrieve from and leveraging the LLM to diversify the types of probing utterances. \jy{double check.} 
Our experiments are in two folds; the first stage is where 49 participants solve the detection tasks individually and the second stage is where participants form a group and solve the questions collectively. We employ a between-subjects design where 10 groups have DeepFakeDeLiBot involved and the remaining does not. We conduct a statistical comparative analysis of detection performance across three different setups (solo vs. group without DeepFakeDeLiBot vs. group with DeepFakeDeLiBot). 

The results of our experiments consistently highlight the superior performance of group problem-solving compared to individual problem-solving. Although the improvement of groups that also interacted with DeepFakeDeLiBot was numerically higher, this improvement was not statistically significant. Groups interacting with DeepFakeDeLiBot exhibited more positive group dynamics, including higher engagement levels, more even participation, better consensus formation, and enhanced probing qualities. Our analysis suggests several conditions and contexts under which DeepFakeDeLiBot promotes performance gains via deliberation, examining this through the lens of participants' backgrounds, group dynamics, and the bot's interaction patterns.

To summarize, our contributions are as follows: (1) We present the first deliberation-enhancing conversational agent specifically designed for deepfake text detection; (2) Our statistical analysis reveals that groups outperform individuals in deepfake text detection. Additionally, groups interacting with DeepFakeDeLiBot demonstrated more positive group dynamics, suggesting that DeepFakeDeLiBot can enhance deliberation effectiveness without compromising detection performance; (3) We explore the effect of a variety of features and the involvement of DeepFakeDeLiBot on performance gain.

\section{Related Work}
%\jy{I'll finish this section on Saturday.}
\subsection{Human Evaluation of Deepfake Text}

As generative models become more capable of producing coherent, contextually appropriate, and convincing text, distinguishing between human and machine-generated content has become challenging. This has led to a growing interest in understanding how human evaluators assess the authenticity and credibility of deepfake text. According to \citet{garbacea-etal-2019-judge}, evaluators could detect reviews generated by Word LSTM and GAN models with 66.61\% accuracy. More recent works such as \citet{ippolito-etal-2020-automatic} and \citet{ippolito-etal-2020-automatic} re-evaluated humans' detection performance on modern LLMs including GPT-2 and GPT-3 and found that their performance was slightly better than random guessing. The authors further attempted to train the evaluators by providing detailed instructions or walking through the task together, but there was a minor performance gain. 

While the majority of prior works have framed the task as a binary classification—determining whether an entire text is generated by humans or machines—\citet{dugan2023real} and \citet{uchendu2023does} are among the first to explore human detection of the transition point where authorship switches from human to machine. Specifically, \citet{dugan2023real} demonstrated that framing the deepfake text detection task as a Real or Fake Text (RoFT) detection game, introduced by \citet{dugan-etal-2020-roft}, enables participants to achieve an accuracy of 72.3\%. In contrast, \citet{uchendu2023does} focused on how collaborative decision-making impacts detection performance. Building on these findings, this study investigates the role of a dialogue agent specialized for driving effective group deliberation in enhancing detection performance and discussion quality.

\subsection{AI-Assisted Group Deliberation}
Research has shown that AI-powered dialogue agents are capable of supporting group decision-making and deliberation across domains without the need for human intervention \cite{sahab2024conversational, kim2021moderator}. By leveraging advanced machine learning and natural language processing techniques, these agents seamlessly drive group discussions by encouraging participation, posing insightful questions, and summarizing key discussion outcomes \cite{agarwal2024conversational, kim2020bot}. One standout example is DeLiBot (Deliberation Enhancing Bot), developed by \citet{karadzhov2023delidata}. Unlike traditional systems, DeLiBot is built to foster constructive group deliberation through strategic probing, often delivered in the form of three (moderation/solution/reasoning) different types of probing questions. Specifically, the bot monitors the dialogue histories, and tracks group dynamics, conversational patterns, and participant interactions to identify optimal moments for intervention. Upon determining the need for intervention, DeLiBot responds with tailored prompts designed to stimulate deeper reflection and improve group performance. The authors suggest that groups engaging with the bot achieved better solutions collectively than individuals could on their own when solving the Wason card selection task. To the best of our knowledge, there is no prior work that investigated how AI-assisted group deliberation affects deepfake detection performance.

%focus on a deliberation-enhancing bot without domain-specific knowledge in order to foster more flexible, collaborative reasoning among participants. This approach encourages critical thinking and diverse perspectives, rather than relying on pre-determined solutions, and allows the group to engage in a more dynamic and open-ended decision-making process.

\section{Methodology}

\begin{figure*}[h!]
    \centering
    \includegraphics[width=0.9\textwidth]{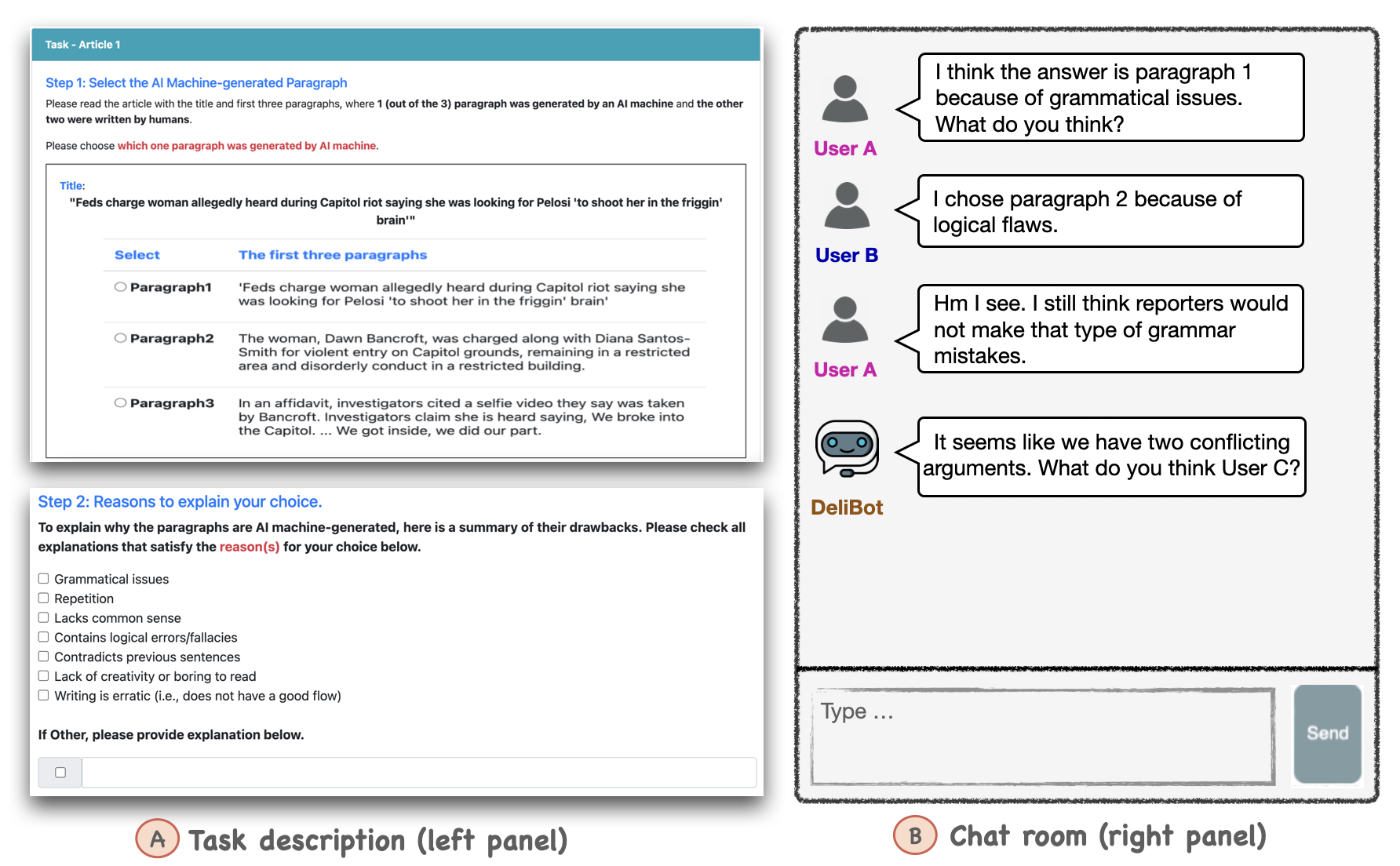}
    \caption{User interface example.}
    \label{fig:interface}
\end{figure*}

%This section first describes how we leverage the existing dataset from prior work \cite{uchendu2023does} to curate more challenging examples using GPT-3.5 for deepfake data curation. We then explain the implementation of DeepFakeDeLiBot and human study design. 

\subsection{Deepfake Data Curation}
Prior literature (e.g., \citet{tulchinskii2024intrinsic}, \citet{mitchell2023detectgpt}, \citet{mireshghallah-etal-2024-smaller}) primarily focused on the detection of sentences or paragraphs solely composed by LLMs. Yet, in the real-world setting, it is more likely that humans employ the generative models to improve the quality of certain parts of their draft \cite{lingard2023writing}. Also, they may amend or replace portions of their written content and evade LLM-text detectors \cite{sadasivan2023can}. Hence, in this study, we perform synthetic data generation where the articles comprise two paragraphs authored by humans and one paragraph generated by the LLM. This design is grounded on \citet{uchendu2023does}, and we started with their dataset. Specifically, the authors selected 50 human-written news articles (mostly from the politics domain) and randomly replaced one out of three paragraphs with artificial texts written by GPT-2. 
We selected news article, particularly in the political domain, because they represent a high-stakes setting where deepfake texts can cause significant societal harm, such as spreading misinformation, shaping public opinion, or destabilizing democratic processes \cite{barari2025political, gosse2020politics}. Moreover, political content tends to elicit stronger reasoning, skepticism, and activation of prior beliefs compared to more neutral domains like weather reports or product reviews \cite{taber2006motivated}, making it a valuable testbed for studying the dynamics of collective judgment.

GPT-2 was released in early 2019 and is estimated to be 100 times smaller than their newer models like GPT-3.5. We hypothesize that texts written by GPT-2 are prone to make writing errors and hence easier to detect from humans' lens than those from larger models due to scaling laws in generation capabilities \cite{kaplan2020scaling}. To ensure that our experiment reflects the current state of LLMs and possesses a good balance in task difficulties, we attempt to regenerate half of the dataset using GPT-3.5 and include them in our experiment.\footnote{At the time of the generation, the most recent models including GPT-4 and GPT-4o were not released.} %We do not replace all of the samples because we do not want to dramatically increase the overall difficulty of the task. 

Our primary goal in dataset curation is to ensure a balanced mix of \textit{easy} and \textit{challenging} questions in the final set.
To gauge the difficulty levels of each article, we leverage three SOTA LLMs (GPT-3.5, LLAMA2-70B-chat, and Claude-2) as judges and feed the questions to the prompt. If all models considered a correct response to the question, we consider it to be an \textit{easy} question. Out of the 50 questions in \citet{uchendu2023does}'s dataset, 16 questions were answered correctly by three LLM evaluators. We consider these 16 questions as \textit{easy} questions. To construct \textit{challenging} articles, We use GPT-3.5 to regenerate the remaining 32 questions. Here we follow the ``fill-in-the-blank" prompting approach, where the model was asked to fill in the empty paragraph slot given the original article title and two human-authored paragraphs. For instance, the prompt template to replace the second paragraph is illustrated below:

\begin{mybox}
Given the title and two paragraphs of news articles, write Paragraph 2 on your own. \\
\newline \textbf{Article Title}:  \texttt{\$\{title\}}
\newline \textbf{Paragraph 1}: \texttt{\$\{paragraph\_1\}}
\newline \textbf{Paragraph 2}: 
\newline \textbf{Paragraph 3}: \texttt{\$\{paragraph\_3\}}
\end{mybox}

\noindent Refer to Table \ref{tab:deepfake-example} for generation examples. After the completion of data curation, we manually inspected GPT-3.5's generation results and filtered out samples lacking consistency and coherence.
%\lee{this needs to be expanded. The quality of generation and how smooth a generated paragraph is among 3 paragraphs are critical. Can we also run some NLP tests to estimate the flowness and quality of generated paragraphs and show that they are good, not easily detectable? I think Adaku did similar tests.}
This resulted in 28 questions out of 32 questions. Additionally, we automatically measured the coherence of three paragraphs as a whole by computing the paragraph-level cosine similarity after encoding the paragraphs with the T5 model \cite{raffel2020exploring} and confirmed that all scores were above 0.8.

\subsection{Implementation of DeepFakeDEliBot}

\begin{table*}[!h]
    \centering
    \footnotesize
    \begin{tcolorbox}[colback=white, colframe=black, title = {In-Context Learning Example}]
    \textbf{Example Context:}\\
    Lion: I chose 3, but looking over it now to see why I chose that\\Zebra: Actually 3 doesnt flow,  reasonable effort to win elecetion as many votes as I could\\Lion: I see why I chose 3. I didn't see where it connected to the 1st 2 paragraphs.\\Zebra: I have to say, in my experience with ChatGPT, I don't see this type of error\\Dolphin: I believe it is paragraph 2\\Zebra: I now think 3 is incorrect as it is not objectively relevant to topic and other paragraphs\\ \\ \textbf{Example Retrieved Probing:} \\ Dolphin, can you please explain a little ? \\
    \\ \textbf{Example Modified Probing:} \\
    Dolphin, can you please explain why you believe it is paragraph 2?
\end{tcolorbox}
\caption{An in-context learning example for GPT-3.5-Turbo to generate synthetic dataset.}
\label{tab:icl-example}
\end{table*}

%\neo{Need a diagram? I can make it more precise if it's not clear. But we may want to say something more, especially about the turn taking in the appendix?}

%\jy{hi Neo, can you please address Andreas's comments here? Also, can you please highlight the key changes we made to the model?}
%\neo{Sure, I've already covered these points in the appendix. I will merge them here. You may need to cut some lines with respect to the page limit though.}

In the original implementation of DeLiBot \cite{karadzhov2024delibots}, probing utterances were retrieved from DeliData \cite{karadzhov2023delidata}, a dataset of group discussions focused on the Wason Card Selection Task \cite{wason1968reasoning}. While DeliData provides general probing utterances, many are overly specific to this task, limiting their applicability to other domains. Additionally, its approach to generation, which involves replacing user and choice mentions in retrieved utterances using mask-filling, restricts flexibility and accuracy. To address these limitations and adapt DeLiBot to our domain, we collected and manually annotated domain-specific data and introduced an additional natural language generation component to produce probing utterances that better reflect the current conversation.

\rparagraph{Inference.} DeepFakeDeLiBot operates as a retrieval-based dialogue agent, maintaining a database of paired dialogue histories and probing utterances. During a conversation, it tracks the dialogue history and retrieves probing utterances through similarity-based retrieval. The retrieval context comprises the 5 utterances preceding the probing utterance, as specified by a hyperparameter. A Sentence-T5 language model computes embeddings for the current context, and cosine similarity is used to compare these embeddings with those in the database \cite{ni2021sentence}. DeepFakeDeLiBot selects the top 5 probing utterances with the most similar contexts and identifies the one expected to yield the greatest performance gain, estimated using an off-the-shelf Tactics-Strategy classifier. The selected utterance is then refined by a fine-tuned Flan-T5 \cite{longpre2023flan} Base language model \cite{chung2024scaling}, generating a probing utterance tailored to the conversation history with accurate participant and choice mentions.

\rparagraph{Training.}
To adapt DeepFakeDeLiBot for the domain of DeepFake debunking, we expanded and refined the dataset of probing utterances, as summarized in Table \ref{tab:dataset_statistics}. First, we manually annotated 5 transcribed group conversations related to the DeepFake task, identifying and extracting relevant probing utterances. These were combined with a filtered subset of DeliData to create an initial dataset. Next, we conducted 10 pilot studies using DeepFakeDeLiBot with this dataset, manually annotating the resulting dialogues to further enrich the context data. Additionally, we leveraged GPT-3.5-Turbo to paraphrase retrieved utterances, augmenting the dataset with more domain-specific probing utterances.

To ensure that retrieved utterances are contextually appropriate, including accurate references to specific users and choices, we fine-tuned a Flan-T5 Base model. This process was framed as a sequence-to-sequence task: given the dialogue context and a retrieved probing utterance, the model generates a modified probing utterance with correct references. For training data creation, we employed 3-shot in-context learning with GPT-3.5-Turbo to generate synthetic pairs of context and retrieved utterances. An example of a manually annotated in-context learning demonstration is provided in Table \ref{tab:icl-example}. The fine-tuned Flan-T5 model was trained over three epochs, achieving strong performance in generating contextually appropriate probing utterances.

\subsection{Human Study Design} 
\rparagraph{Participant Recruitment.} We used Upwork\footnote{\url{https://www.upwork.com}}, one of the largest freelance websites that has skilled freelancers in diverse domains such as writing, design, and web development. To promote our experiment on Upwork, we registered as a client and posted our research objectives and task descriptions. We explicitly mentioned that this posting is for research purposes and attached the consent form to the job posting for review. In addition, the following requirements were highlighted in the post: (1) participants should be at least 18 years old and (2) participants should be fluent in English. All freelancers could view our job posting, and those who were willing to participate were asked to return a questionnaire. It included three questions we crafted to understand their backgrounds better: 
(1) What is the highest level of degree you have completed in school?; (2) Did you major in English or English Literature?; and (3) Describe your recent experience with similar projects. 
Once we reached a reasonable number of applications, we verified participants' eligibility by checking their self-reported age, language, and education in the profile. We also examined their desired hourly wage, as it substantially varied (ranging from \$15 to \$100) depending on expertise and experience. To limit the influence of desired payment differences on the experiment results, we only hired participants requesting \$30-\$35 per hour, which resulted in a total of 49 participants. Lastly, we collected the signed consent form and activated the contracts. The contracts were required by default in the Upwork platform to guarantee that freelancers and clients agree upon clients' requested pay rate and that clients compensate freelancers based on submitted hours through the Upwork system. Assuming an average task duration of three hours per participant, the estimated total compensation for all participants was \$7,350 USD. 

\rparagraph{Experiment Design.} Our study consists of two stages with two pre- and post-study surveys. Each stage of our experiment contains 14\footnote{To avoid participant cognitive fatigue, we reduced the total number of questions by sampling 7 questions from GPT-2 generated articles and 7 questions from GPT-3.5 generated articles.} articles. In the first stage, we ask 49 participants to solve the deepfake detection task on their own. Upon completion, they are redirected to the pre-study survey link where we ask questions related to their backgrounds and self-perceived performance (see Table \ref{tab:survey}). The second stage, on the other hand, gathers groups of randomly selected individuals (consisting of two to three people) and asks them to discuss their answers to the same questions from the first experiment. This resulted in 20 groups. We then randomly assigned half of the groups (n=10) to engage with DeepFakeDeLiBot. Among 49 participants, we have 25 participants in the DeepFakeDeLiBot setup, and the remaining 24 participants were asked to solve the questions without  DeepFakeDeLiBot. 

To support the synchronous discussion, we included a chat service in our web interface. All members are assigned anonymized user names and can freely discuss their choices and reasoning. We explicitly informed participants that they were free to submit their own individual responses, regardless of whether the group reached a consensus. Lastly, participants were prompted to complete the post-study survey that inquires about participants' experiences with group collaboration and interaction with DeepFakeDeLiBot (see Table \ref{tab:survey}).

%Specifically, the post-study survey inquires participants' experiences with group collaboration and overall performance as a group. If their experiment included DeepFakeDeLiBot, additional questions were asked such as usefulness of DeepFakeDeLiBot and feedback for improvement. All questions are reported in Appendix for reference. 

%All communication not related to the actual experiment took part in Upwork's  messaging system. We used Qualtrics\footnote{\url{https://www.qualtrics.com}} service to generate the survey forms. According to our mock experiment, it took roughly 2 minutes to complete one question individually. In the group setting, it was extended to 6 minutes per question. To avoid participant cognitive fatigue, we utilize 7 questions from the "easy" question pool and 7 questions from the "hard" question pool as a final set. Expected time for participants to complete all stages of the experiment was 1.5-2 hours. 

\section{Results}

%In this section, we answer the following three research questions:
%\begin{itemize}
%    \item RQ1: Does deepfake detection performance differ between individual and collaborative problem-solving, with or without the deliberation-enhancing chatbot?
%    \item RQ2: How does the involvement of a deliberation-enhancing bot affect collaboration dynamics (e.g., engagement, even participation, consensus formation, probing dynamics and change of minds)?
%    \item RQ3: What conditions make group collaboration with DeepFakeDeLiBot effective?
%\end{itemize}

\subsection{RQ1: Deepfake Detection Performance Comparison}

\begin{table}[!t]
\centering
\footnotesize
   \begin{tabular}{@{}c|c|c@{}}
\toprule
\textbf{Setting}                                                                                         & \textbf{Mean Accuracy}                                                             & \textit{\textbf{p}}           \\ \midrule
\begin{tabular}[c]{@{}c@{}}Individual vs. Group \\ (w/o DeepFakeDeLiBot)\end{tabular}                    & \begin{tabular}[c]{@{}c@{}}45.83\% vs. 54.76\%\\ (8.93\% $\uparrow$)\end{tabular}  & {\color[HTML]{FE0000} 0.0004} \\
\begin{tabular}[c]{@{}c@{}}Individual vs. Group \\ (w. DeepFakeDeLiBot)\end{tabular}                     & \begin{tabular}[c]{@{}c@{}}48.86\% vs. 57.43\%\\ (8.67\% $\uparrow$)\end{tabular}  & {\color[HTML]{FE0000} 0.0013} \\
\begin{tabular}[c]{@{}c@{}}Group (w/o DeepFakeDeLiBot) \\ vs. \\ Group (w. DeepFakeDeLiBot)\end{tabular} & \begin{tabular}[c]{@{}c@{}}54.76\% vs. 57.43\% \\ (2.67\% $\uparrow$)\end{tabular} & 0.482                         \\ \bottomrule
\end{tabular}
    %\captionsetup{justification=centering} % Center the caption
    \caption{T-test results for individual detection performance.}
    \label{tab:rq1}
\end{table}

\begin{comment}
\begin{table}[!t]
\centering
\small
    \centering
    \begin{adjustbox}{max width=\columnwidth}
    \begin{tabular}{c|c|c}
    \toprule
    {\textbf{Setting}}     &    {\textbf{Mean Accuracy}}                &  {\textbf{P-Value}}  \\                                                
                                  \midrule \midrule
    Individual vs. Group (w/o DeepFakeDeLiBot)      & \multicolumn{1}{c|}{36.42\% vs. \textbf{51.42\%} (15\% $\uparrow$)}    &  3.39e-05 \\ \cmidrule{1-3}
    Individual vs. Group (w. DeepFakeDeLiBot)   & \multicolumn{1}{c|}{40\% vs. \textbf{57.14\%} (17.14\% $\uparrow$)}    &  1.03e-06 \\ \cmidrule{1-3}
    Group (w/o DeepFakeDeLiBot) vs. Group (w. DeepFakeDeLiBot) & \multicolumn{1}{c|}{51.42\% vs. \textbf{57.14\%} (5.72\% $\uparrow$)} &  0.195    \\ \bottomrule
    \end{tabular}
    \end{adjustbox}
    %\captionsetup{justification=centering} % Center the caption
    \caption{T-test results for detection performance with majority voting. \jy{will regenerate this.}}
    \label{tab:rq1_maj}
\end{table}
\end{comment}

To perform the analysis in a fine-grained manner, we measure the correctness of their submitted responses at the article level and use it for statistical testing. Let's say $C_i$ is the correctness of the response for the $i-th$ article where $C_i \in {0,1}$, with 1 indicating a correct response and 0 indicating an incorrect response.  For a given user, the responses are represented as $[C_1, C_2, C_3, ..., C_{14}]$. We then compute the statistical power of their mean differences before and after group collaboration through a paired t-test. When comparing groups with or without DeepFakeDeLiBot, we instead perform the unpaired t-test since the two populations (groups without DeepFakeDeLiBot vs. groups with DeepFakeDeLiBot) are independent. 

Table~\ref{tab:rq1} reports mean performance differences and their statistical significance. The results show that collaborative problem-solving consistently achieves higher performance in deepfake text detection than independent problem-solving by 8\%. Moreover, groups that interacted with the deliberation-enhancing bot achieved the highest detection performance. However, the observed performance difference was not found to be statistically significant compared to the groups without the bot. 

\begin{comment}
We also measured the detection performance via majority voting within the group to better understand how the group as a whole performs when their collective judgment is aggregated. Although the first stage did not include the collaboration components, we compute their majority voting results and use it as the baseline. As shown in Table~\ref{tab:rq1_maj}, even with the majority voting, group settings with and without DeepFakeDeLiBot outperform their solo performance. Since the baseline individual performance has gone lower (e.g., 45.83\% $\rightarrow$ 36.42\%), the degree of performance improvement also increased from 8\% to 16 to 17\%. The comparison between group experiments with and without DeepFakeDeLiBot shows that the group performance as a whole tended to be higher with the involvement of DeepFakeDeLiBot as opposed to without DeepFakeDeLiBot by 5.72\%. Yet, the difference was found to be statistically insignificant.
\end{comment}

%We notice there is a performance gap in individuals' baseline performance (45.83\% vs. 48.86\%). This may have affected our result interpretation, as it is more challenging to further see the gain on top of already superior performance. Hence, we run another comparative analysis by dividing groups into two subgroups with comparative initial performance. 

\subsection{RQ2: DeepFakeDeLiBot and Collaboration Dynamics}

% \usepackage{booktabs}
% \usepackage{multirow}
\begin{table*}[]
\centering
\footnotesize
\begin{tabular}{@{}c|ccc|cc@{}}
\toprule
                                    & \multicolumn{3}{c|}{\textbf{T-Test}}                                                                                                                                                                                                  & \multicolumn{2}{c}{\textbf{Linear Regression}}                    \\ \cmidrule(l){2-6} 
\multirow{-2}{*}{\textbf{Features}} & \multicolumn{1}{c|}{\textbf{\begin{tabular}[c]{@{}c@{}}Group \\ w/o DeepFakeDeLiBot\end{tabular}}} & \multicolumn{1}{c|}{\textbf{\begin{tabular}[c]{@{}c@{}}Group \\ w. DeepFakeDeLiBot\end{tabular}}} & \textit{\textbf{p}}          & \multicolumn{1}{c|}{\textbf{Coef}} & \textit{\textbf{p}}          \\ \midrule
Participant engagement              & \multicolumn{1}{c|}{4.92}                                                                          & \multicolumn{1}{c|}{6.15}                                                                         & {\color[HTML]{FE0000} 0.006} & \multicolumn{1}{c|}{-0.01}         & 0.17                         \\
Even participation                  & \multicolumn{1}{c|}{0.04}                                                                          & \multicolumn{1}{c|}{0.03}                                                                         & 0.26                         & \multicolumn{1}{c|}{-0.01}         & 0.97                         \\
Consensus Formation                 & \multicolumn{1}{c|}{0.85}                                                                          & \multicolumn{1}{c|}{0.95}                                                                         & {\color[HTML]{FE0000} 0.005} & \multicolumn{1}{c|}{0.19}          & {\color[HTML]{FE0000} 0.001} \\
Solution Probing Frequency          & \multicolumn{1}{c|}{1.44}                                                                          & \multicolumn{1}{c|}{1.04}                                                                         & 0.34                         & \multicolumn{1}{c|}{1.05}          & 0.19                         \\
Reasoning Probing Frequency         & \multicolumn{1}{c|}{1.68}                                                                          & \multicolumn{1}{c|}{3.48}                                                                         & {\color[HTML]{FE0000} 0.001} & \multicolumn{1}{c|}{-0.48}         & 0.44                         \\
Moderation Probing Frequency        & \multicolumn{1}{c|}{6.51}                                                                          & \multicolumn{1}{c|}{7.26}                                                                         & 0.47                         & \multicolumn{1}{c|}{0.08}          & 0.93                         \\
Diversity of Discussed Solution     & \multicolumn{1}{c|}{0.82}                                                                          & \multicolumn{1}{c|}{1.05}                                                                         & {\color[HTML]{FE0000} 0.02}  & \multicolumn{1}{c|}{0.10}          & {\color[HTML]{FE0000} 0.04}  \\
Diversity of Submitted Reasoning    & \multicolumn{1}{c|}{2.76}                                                                          & \multicolumn{1}{c|}{2.75}                                                                         & 0.92                         & \multicolumn{1}{c|}{0.06}          & {\color[HTML]{FE0000} 0.008} \\ \bottomrule
\end{tabular}
\caption{T-test results for collaboration dynamic comparison w.r.t. DeepFakeDeLiBot usage and linear regression results of collaboration dynamics and performance gain.}
\label{tab:rq2}
\end{table*}

In our experiments, we assessed collaboration dynamics by utilizing engagement levels, even participation, consensus formation, discussion constructiveness, and probing utterance frequencies as key proxies. These features were computed based on exchanged utterances between group members and submitted responses. Specifically, we tracked the following items at the article level: participant engagement, even participation, consensus formation, the frequency of solution-driven probing utterances, the frequency of reasoning-driven probing utterances, the frequency of moderation-driven probing utterances, the diversity of discussed solutions and reasoning. Their definitions and computational measurement approaches can be described as follows:
\begin{itemize}
    \item \textbf{Participant engagement}: we measure the participants' engagement levels in the discussion by computing the average number of utterances spoken per participant.
    \item  \textbf{Even participation}: we measure whether group members contributed evenly to the discussion (not dominated by particular individuals) by computing the variance in the distribution of participants' engagement rate.
    \item \textbf{Consensus formation}:  we measure whether participants successfully formed a consensus by examining their submitted responses. If all group members submitted an identical response, we assign 1. Else, we assign 0. 
    \item \textbf{Frequency of probing utterances}: we measure how often participants leveraged probing utterances to drive discussion by computing the percentage of three types of probing utterances (moderation (e.g., \textit{Let’s discuss our initial solutions}), solution (e.g., \textit{Why did you think it wasn’t paragraph 3?}, reasoning (e.g., \textit{Are we going for paragraph 2?}) of all utterances. Here we use the \citet{karadzhov2023delidata}'s finetuned classification model from DeliToolkit\footnote{\url{https://github.com/gkaradzhov/delitoolkit}}.  
    \item \textbf{Diversity of discussed solutions}: we measure how often changes of mind occur by tracking the participants' solution mentions within the dialogue. To achieve this, we utilized a regular expression to extract their mentions of paragraphs from their utterances. 
    \item \textbf{Diversity of submitted reasoning}: we aim to measure whether participants exchanged various justification categories throughout the discussion by counting the unique number of submitted reasoning types. 
\end{itemize}

\noindent For this analysis, we excluded utterances generated by DeepFakeDeLiBot. We performed the unpaired t-test to validate whether the mean differences of these features between the two conditions (groups without DeepFakeDeLiBot vs. groups with DeepFakeDeLiBot) were statistically meaningful.

As shown in Table~\ref{tab:rq2}, the differences in participants' engagement levels, consensus formation, the frequency of reasoning probing utterances, and the diversity of solutions considered were found to be statistically significant. In particular, groups that engaged with DeepFakeDeLiBot tended to have 1.25\% higher interaction rates with their team members than groups without DeepFakeDeLiBot. The probability of reaching a consensus was also higher for groups with DeepFakeDeLiBot. Among the three types of probing utterances, participants from the Delbiot group exchanged reasoning-probing utterances more often than participants in groups without DeepFakeDeLiBot. The frequency of moderation-probing utterances was slightly higher for the groups with DeepFakeDeLiBot compared to the groups without DeepFakeDeLiBot. Yet, the differences were statistically insignificant. Decision-making processes were measured through discussed solutions and submitted reasoning. Although there was no meaningful difference in the number of submitted reasoning between the two groups, groups with DeepFakeDeLiBot were more likely to discuss more diverse solutions than groups without DeepFakeDeLiBot.   

%discuss how these roles affect the group's performance
These findings naturally led us to the next question: \textit{what is the relationship between collaboration dynamics and performance as a whole?}. To answer this question, we ran a linear regression model by setting 8 features as independent variables and group performance as a dependent variable. The dependent variable was represented by the percentage of group members answering the question correctly during the group session. For example, if two out of three people within the group submitted an accurate response, their performance as a whole is equivalent to 66.66\%.  
Table \ref{tab:rq2} demonstrates the linear regression results. The model suggests that consensus formation, the diversity of discussed solutions, and submitted reasoning are strong predictors of performance gain. The coefficient of 0.1981 indicates that for every one-unit increase in consensus formation, the performance gain is expected to increase by 19.81\% units, holding other factors constant. Similarly, the diversity of discussed solution and submitted reasoning have positive relationships with performance gain; for every one-unit increase in the diversity of discussed solution and reasoning, the performance gain is expected to increase by 10.5\% and 6\% units, respectively.

\subsection{RQ3: What conditions make group deliberation with DeepFakeDeLiBot effective?}
This RQ is to identify the conditions or contexts where DeepFakeDeLiBot has the greatest positive impact. Factors such as participants' profiles, the number of group members, user interaction patterns, or the type of utterances provided by DeepFakeDeLiBot can impact the degree of performance boost resulting from the bot.

\rparagraph{Participants' backgrounds and experiences.} We first analyzed how participants' self-reported characteristics submitted through pre- and post-study surveys influence the performance gap. Specifically, we examine their self-perceived proficiency in writing (Q7), AI-powered tool usage levels (Q9), and their trust levels in AI-powered tools (Q10) from the pre-study survey. In addition, their self-perceived performance after group collaboration (Q2) and self-perceived effectiveness of group collaboration (Q3) from the post-study survey were studied. The performance gap was computed by subtracting the solo performance from the group performance. If the particular individual answered more questions correctly after the group session, the performance gap would be a positive value. If not, it would remain 0 or negative. To investigate the effect of these variables and chatbot usage on the performance gap, we run a linear regression. We also modeled interaction effects between independent variables in order to reveal whether the impact of DeepFakeDeLiBot is stronger, weaker, or even reversed under certain conditions. Our results (Table \ref{tab:rq3_results}) indicate that neither the chatbot nor the individual predictors (Q7, Q9, etc.) have significant main effects. Yet, we find a positive and significant coefficient regarding interaction terms between Q3 and the involvement of DeepFakeDeLiBot (coefficient= 9.6432, \textit{p} = 0.034). Overall, This suggests that for individuals who perceive group collaboration as effective, the chatbot adds substantial value to their performance.

\rparagraph{Group Dynamics.} Next, we examine the effect of features related to group dynamics (from RQ2) on the performance gain before and after the group discussion. As these features such as participant engagement, even participation, and consensus formation are calculated at the article level, we calculate the performance gap at the article level as well. This can be done by measuring the gap in the percentage of people who answered the question correctly. For instance, in article 1, if one out of three (i.e., 33.33\%) participants submitted the correct response in the solo session and then everyone (i.e., 100\%) submitted the correct response after the group session, the performance gap is 66.67\%.  We also modeled interaction effects between independent variables to reveal whether the impact of DeepFakeDeLiBot is stronger, weaker, or even reversed under certain conditions. Our results (Table \ref{tab:rq3_result2}) show that neither DeepFakeDeLiBot nor the individual group dynamic predictors have significant main effects on the performance. Yet, we find a negative and significant coefficient regarding interaction terms between the moderation probing utterance frequency and the involvement of DeepFakeDeLiBot (coefficient = -0.9091, \textit{p} = 0.05). For each unit increase in moderation probing utterance percentage, the effect of chatbot involvement on performance gain decreases by -0.9091 units. Alternatively, it means the relationship between chatbot involvement and performance gain becomes more negative as the user-generated moderation probing utterance percentage increases.

\rparagraph{Participants' interaction with DeepFakeDeLiBot.} We further delve into why certain groups achieve a noticeable performance boost among 10 groups that engaged with DeepFakeDeLiBot. Among 10 groups, 6 groups outperformed their initial solo performance. The remaining group demonstrated no gain or even regress from their original detection performance. We speculate that their interaction patterns with DeepFakeDeLiBot will differ between these two subsets. To validate this assumption, we investigate 5 various categories:
\begin{itemize}
    \item  \textbf{DeepFakeDeLiBot's engagement rates} : we measure DeepFakeDeLiBot's engagement levels in discussion by computing the percentage of DeepFakeDeLiBot's utterances out of all participants' utterances.
    \item \textbf{Frequency of DeepFakeDeLiBot-generated probing utterances} : we quantify how often DeepFakeDeLiBot generated three (moderation, solution, reasoning)) types of probing utterances by computing the percentage of each type across DeepFakeDeLiBot's responses.
    \item \textbf{Participants' unresponsiveness to DeepFakeDeLiBot's responses} : we measure how often participants ignored DeepFakeDeLiBot's responses by modeling the conversation flow. Motivated by \citet{uchendu2023does}, we leverage the pre-trained LLMs\footnote{Since the authors had to run the model in the local machine, we used the GPT-2 small model.}' perplexity\footnote{Perplexity quantifies how well a language model predicts the next word in a sequence. Lower perplexity indicates better predictive confidence, which can signal whether the transition between two sentences flows naturally or not.} scores. Since our objective is to identify if participants respond to DeepFakeDeLiBot's utterances or not, we can take DeepFakeDeLiBot's utterance as the first sentence and consider the following sentence as the participants' response. For measurement, we first take the DeepFakeDeLiBot's utterance as baseline perplexity and check if the combined sequence's perplexity is lower than the baseline. 
    \item \textbf {Lexical diversity of DeepFakeDeLiBot's utterances} : we measure the lexical diversity of DeepFakeDeLiBot's responses by counting the number of unique n-grams.
    \item \textbf{Semantic coherence of DeepFakeDeLiBot' utterances} : we measure how semantically coherent DeepFakeDeLiBot's responses are to the ongoing conversation by calculating the cosine similarity between the embedding vectors of previous turns and DeepFakeDeLiBot's response. For text encoding, we use the T5 model.
\end{itemize}

\noindent For statistical analysis, we first performed the t-test to test whether the means of these factors were significantly different between two subgroups (a group without performance gain vs. a group with performance gain).

As illustrated in Table \ref{tab:rq3}, the means of DeepFakeDeLiBot's engagement rates, its generation frequency of moderation-probing utterances, and lexical diversity are found to be statistically different. For instance, while DeepFakeDeLiBot's participation rate was 8\% on average for groups with a performance gain, DeepFakeDeLiBot engaged less often (5\%) within groups that did not have the performance boost. Additionally, DeepFakeDeLiBot more frequently generated moderation-driven probing utterances in groups that experienced performance gain compared to those without any gain. Reasoning-probing utterances were less frequent in groups with performance gain, but the difference was not statistically significant. In comparing dialogue quality through the lens of lexical diversity and semantic coherence, only the lexical diversity of DeepFakeDeLiBot tended to be higher within groups with performance gain than groups without performance gain. Unresponsiveness rates were moderately high (41\%-44\%) in both groups, but the mean difference was not statistically meaningful.

Subsequently, we conducted a linear regression analysis to determine whether these factors could predict performance gain before and after collaborative problem-solving with DeepFakeDeLiBot. Interestingly, despite observed differences in means, none of these features were found to be significant predictors of the measured performance gain.

\section{Qualitative Discourse Analysis}
To better understand why the chatbot’s interventions were ineffective for certain groups, we employ an inductive qualitative approach, allowing codes and themes to emerge organically from the dialogue transcripts. The lead author iteratively reviewed the data and identified recurring patterns, which were then grouped into higher-order categories reflecting group responsiveness to DeepFakeDeLiBot. See Table \ref{tab:code} for the codebook and dialogue examples. Our qualitative analysis suggests that this outcome stems from four core limitations.

\vspace{1mm}
\noindent \textbf{DeepFakeDeLiBot's temporal misalignment with conversation flow:} In many low-performing groups, DeepFakeDeLiBot’s prompts were poorly timed relative to the group's conversational flow or decision points. Rather than entering at moments of uncertainty or open debate, the bot often interjected after participants had already reached consensus or shifted to a different topic. In such cases, the bot’s input was perceived as irrelevant or even disruptive to the group’s flow. This misalignment diminished the influence of DeepFakeDeLiBot on group reasoning. %Instead of scaffolding decision-making, the bot’s contributions arrived too late to shape the outcome, revealing a critical limitation in its responsiveness to group dynamics. This finding underscores the importance of interactional timing in human-AI collaboration: without temporal sensitivity, even well-formed prompts may fail to influence group reasoning.

\vspace{1mm}
\noindent \textbf{Participants' lack of receptiveness:} Low-performing groups frequently exhibited a high rate of dismissal toward DeepFakeDeLiBot’s prompts. Even when the bot issued reasonable, well-formed questions, participants often showed low contextual receptiveness and seldom treated the bot as a legitimate contributor to the dialogue. Without active uptake and collaborative elaboration, even well-crafted prompts had minimal downstream impact on collective reasoning. This pattern suggests that Delibot’s effectiveness is shaped not only by the quality of its interventions, but also by the group’s disposition toward external input and its overall orientation toward deliberative engagement.

\vspace{1mm}
\noindent \textbf{Participants' accelerated task completion:} In some groups, participants moved through the task so quickly that Delibot had little opportunity to intervene. These groups prioritized rapid consensus over deliberative reasoning, often finalizing decisions before the bot's prompts could appear or be meaningfully engaged with. As a result, even well-timed or contextually relevant interventions were rendered ineffective—not because of their content, but because the conversational window had already closed. This dynamic reveals a structural mismatch between the pace of group interaction and the bot’s ability to scaffold reasoning in real time.

\section{Discussion and Conclusion}
%We elaborate on our key findings regarding three RQs in this section. 

\rparagraph{RQ1. While group collaboration improved deepfake detection performance, the involvement of the deliberation-enhancing bot did not further enhance performance.}

%\noindent Consistent with well-established findings in collective intelligence, we observe that group deliberation could substantially enhance deepfake detection performance compared to individual efforts. This improvement is likely to stem from the ability of groups to exchange diverse perspectives and critically evaluate the information \cite{fleenor2006wisdom}. According to the exit survey, participants mentioned that group collaboration gave them a second pair of eyes with an individual of a different professional background and gave them more confidence in their responses. Moreover, collaborative problem solving helped them identify unusual patterns in deepfake content that might be overlooked by individuals working in isolation. Lastly, the interactive nature of deliberation encouraged iterative refinement of judgments, as participants challenge each other’s assumptions and reach the best final answer. Our results indicate that the inclusion of the deliberation-enhancing bot (DeepFakeDeLiBot) helps participants maintain their improved group performance. When group performance was evaluated using majority voting, groups with DeepFakeDeLiBot slightly outperformed those without it in terms of performance gain. Yet, its statistical power was limited. This finding contradicts our expectations, as deliberation is a key to collective decision making and is typically associated with improved group outcomes \citet{karadzhov2023delidata, iaryczower2018can}. 

\noindent Our findings that group deliberation significantly enhances deepfake detection performance compared to individual efforts. This improvement likely arises from the exchange of diverse perspectives and the ability to critically evaluate information \cite{fleenor2006wisdom}. According to the exit survey, participants noted that group collaboration provided a ``second pair of eyes" from individuals with different professional backgrounds, boosting their confidence in responses. Moreover, working together helped them identify unusual patterns in deepfake content that might be missed by individuals working alone. %The iterative nature of deliberation also encouraged participants to challenge assumptions and refine their judgments, leading to better final answers. Our results suggest that the deliberation-enhancing bot (DeepFakeDeLiBot) helped maintain group performance. 

While groups with DeepFakeDeLiBot slightly outperformed those without it in terms of performance gain, the effect was not statistically significant. This finding challenges our expectations, as deliberation is typically linked to improved group outcomes in decision-making \cite{karadzhov2023delidata, iaryczower2018can}. There are two possible explanations for our findings. First, some participants in the exit survey noted that DeepFakeDeLiBot felt redundant, as the group was already effectively using its collective knowledge and reasoning. This minimized reliance on the bot, reducing its impact on the discussion. Second, deepfake detection may require specialized skills beyond what group deliberation alone can offer. %Despite balancing task difficulty, participants' initial accuracy was low—68\% for GPT-2 generated paragraphs and just 25\% for GPT-3, highlighting the challenge of detecting texts generated by modern LLMs. While group deliberation is useful, tasks like deepfake text detection may require more specialized knowledge. 
\citet{uchendu2023does} found that experts outperformed laypeople in detecting deepfake texts, suggesting that advanced linguistic skills matter. Future research should explore models that provide task-specific cues while supporting deliberation based on the group's progress.

\begin{table*}[]
\centering
\footnotesize
\begin{tabular}{@{}c|ccc|cc@{}}
\toprule
                             & \multicolumn{3}{c|}{\textbf{T-Test}}                                                                                                                                                                                                    & \multicolumn{2}{c}{\textbf{Linear Regression}}           \\ \midrule
\textbf{Features}            & \multicolumn{1}{c|}{\textbf{\begin{tabular}[c]{@{}c@{}}Group w/o \\ Performance Gain\end{tabular}}} & \multicolumn{1}{c|}{\textbf{\begin{tabular}[c]{@{}c@{}}Group w.\\ Performance Gain\end{tabular}}} & \textit{\textbf{p}}           & \multicolumn{1}{c|}{\textbf{Coef}} & \textit{\textbf{p}} \\ \midrule
Engagement Rates             & \multicolumn{1}{c|}{0.05}                                                                           & \multicolumn{1}{c|}{0.08}                                                                         & {\color[HTML]{FE0000} 0.01}   & \multicolumn{1}{c|}{0.69}          & 0.13                \\
Solution Probing Frequency   & \multicolumn{1}{c|}{0.0}                                                                            & \multicolumn{1}{c|}{0.01}                                                                         & 0.27                          & \multicolumn{1}{c|}{0.48}          & 0.14                \\
Reasoning Probing Frequency  & \multicolumn{1}{c|}{0.23}                                                                           & \multicolumn{1}{c|}{0.16}                                                                         & 0.27                          & \multicolumn{1}{c|}{-0.08}         & 0.55                \\
Moderation Probing Frequency & \multicolumn{1}{c|}{0.26}                                                                           & \multicolumn{1}{c|}{0.55}                                                                         & {\color[HTML]{FE0000} 0.0001} & \multicolumn{1}{c|}{-0.08}         & 0.48                \\
Lexical Diversity            & \multicolumn{1}{c|}{0.44}                                                                           & \multicolumn{1}{c|}{0.68}                                                                         & {\color[HTML]{FE0000} 0.0001} & \multicolumn{1}{c|}{0.008}         & 0.95               \\
Semantic Coherence           & \multicolumn{1}{c|}{0.58}                                                                           & \multicolumn{1}{c|}{0.57}                                                                         & 0.69                          & \multicolumn{1}{c|}{0.69}          & 0.13                \\
Unresponsiveness Rates       & \multicolumn{1}{c|}{0.41}                                                                           & \multicolumn{1}{c|}{0.44}                                                                         & 0.12                          & \multicolumn{1}{c|}{0.22}          & 0.17                \\ \bottomrule
\end{tabular}
\caption{T-test results for interaction patterns of DeepFakeDeLiBot w.r.t. group performance gain and linear regression results.} %\jy{will double-check the numbers and update the table}
\label{tab:rq3}
\end{table*}

\rparagraph{RQ2. The deliberation-enhancing bot positively influences collaboration dynamics that are important factors in achieving high performance. }

\noindent Although performance gain between groups with and without DeepFakeDeLiBot did not differ significantly, we hypothesize that DeepFakeDeLiBot may bring secondary benefits, particularly in enhancing collaboration dynamics within the group. Through comparative analyses, we observed that groups utilizing DeepFakeDeLiBot exhibited higher rates of participant engagement, increased consensus formation, more frequent usage of reasoning-probing utterances, and greater diversity in the solutions discussed during collaborative tasks, compared to groups that did not have DeepFakeDeLiBot. Similar findings have been reported in studies in AI-assisted group decision-making processes \cite{kim2021moderator, shin2022chatbots, dortheimer2024evaluating},  where chatbots can foster more active and constructive dialogue.

 Among these factors, consensus formation, the diversity of discussed solutions, and submitted responses emerged as critical predictors of group performance. Specifically, an increase in the strength of consensus was positively correlated with improved group performance. The importance of consensus building has been suggested in the strategic decision-making
procedure \cite{dess1987environment, whyte1989groupthink}. Furthermore, it could promote a more deliberative process \cite{hare1980consensus} and can reduce the probability of errors, producing higher quality decisions \cite{davis1993quantitative, feddersen1998convicting}. Our finding is well-aligned with existing literature. We also found that a higher diversity of solutions and justification types submitted by participants was associated with increased group performance. This result aligns with the previous studies (e.g., \citet{post2009capitalizing}, \citet{hundschell2022effects}) such that diversity of thought stimulates innovation and problem-solving by encouraging the exploration of a broader range of possibilities. %When participants bring varied perspectives and reasoning to the table, it enriches the collective understanding of the problem space and allows the group to evaluate multiple approaches more critically.

Contrary to our initial expectations, neither the overall levels of participant engagement nor the distribution of engagement among group members demonstrated significant relationships with group performance. While engagement is often viewed as a key driver of group effectiveness \cite{yoerger2015participate}, our findings challenge this assumption in the context of the deepfake detection task. They suggest that simply increasing engagement levels or ensuring equitable participation may not be sufficient to improve performance. Similarly, the generation of probing utterances, often expected to enhance group outcomes by stimulating deeper analysis, did not demonstrate a significant impact on group performance. 

%Overall, these observations indicate that DeepFakeDeLiBot, while not directly boosting performance, may enhance the collaborative process in ways that are not immediately measurable by task performance alone.

%Among these factors, consensus formation and the diversity of discussed solutions demonstrated statistically significant positive associations with overall group performance. Specifically, an increase in consensus formation and the diversity of solutions considered within group discussions was consistently linked to improvements in performance metrics. Furthermore, we identified that greater diversity in the reasoning submitted by participants had a potential to enhance group outcomes by fostering more innovative and comprehensive solutions.

%However, contrary to our initial expectations, neither the overall levels of participant engagement nor the distribution of engagement across group members showed significant relationships with group performance. Similarly, the frequency of reasoning-probing utterances, while indicative of deeper cognitive processing and critical evaluation within the group, did not appear to have a measurable impact on overall performance outcomes.

%These findings suggest that while DeepFakeDeLiBot may not directly influence detection performance, its ability to facilitate critical collaborative processes—such as consensus-building and the consideration of diverse solutions—could indirectly contribute to enhancing group effectiveness in complex problem-solving scenarios.

%\vspace{0.2cm}

\rparagraph{RQ3. Participants' backgrounds, experiences, and interaction patterns differ significantly based on performance gain.}

\noindent Several studies \cite{liu2010students, huang2018background} argued that group composition (e.g., the members' backgrounds and beliefs) and group interaction patterns shape the success of the group decision-making. For example, in the context of deepfake text detection, prior literature such as \citet{clark-etal-2021-thats,uchendu2023does} reported that experts in writing domains could identify deepfake texts more accurately than laypeople, as they are more knowledgeable to capture unnatural flow or subtle topical changes often found in deepfake texts. According to our analyses, participants' background information, including their proficiency in writing, experiences in AI tools, and their trust in these tools, were not significant predictors of performance gain. We also examined relationships between their self-perceived detection performance and the effectiveness of group collaboration and performance gain. Our result reveals that, for individuals who perceived group collaboration as effective, the involvement of DeepFakeDeLiBot can positively impact their performance. %This showcases a critical role of participants' overall perception of collaboration in the amount of impact DeepFakeDeLiBot can bring to the group. 
This indicates that, in groups where collaboration is already perceived as smooth and effective, DeepFakeDeLiBot can further amplify these dynamics to achieve even greater performance outcomes. However, in groups where collaboration was deemed unhelpful, DeepFakeDeLiBot's impact may be shadowed by group-specific challenges. 

We empirically validated that, depending on the degree of performance gain, DeepFakeDeLiBot’s engagement rates, its generation frequency of moderation-probing utterances, and lexical diversity differed significantly. Still, these factors did not strongly correlate with predicting performance improvements. 
Our findings in the discourse analysis suggests that the effectiveness of the bot is not uniform, but contingent on participants’ contextual receptiveness and the bot's interjection timing. In groups that were already deliberating effectively, Delibot’s prompts were more likely to be integrated and acted upon.

%This indicates that other, yet unidentified, confounding factors may play a critical role in driving the bot's ability to enhance its performance. Therefore, future research should explore additional potential components such as user feedback and contextual factors (e.g., content type, task difficulty). 

%In predicting the performance gain with group dynamics and the involvement of DeepFakeDeLiBot, group dynamic features, except for the frequency of participants' moderation probing utterance exchange, showed no significant relationships to the performance gain when DeepFakeDeLiBot is involved. The increase in the frequency of moderation probing utterances generated by participants, on the other hand, were likely to reduce the performance gain when DeepFakeDeLiBot is involved. This may be explained by the fact that ...

\section{Limitations}
Our study has two limitations to consider. First, our sample size is relatively small, primarily due to difficulties in recruiting qualified individuals on the Upwork platform and the associated costs of compensation. Moreover, the experiment required synchronous communication, further constraining the sample size due to practical considerations such as participant availability and scheduling. These factors may have limited the statistical power of our analyses. However, given the exploratory nature of this research, our primary aim was to identify trends and relationships rather than to generalize findings to a larger population. Future studies with larger samples can help validate and extend these findings. Second, the generalizability of our findings regarding detection performance may be limited, as we exclusively used two of OpenAI's LLMs to generate deepfake paragraphs. Future research should examine how a broader range of model families, such as Meta's Llama or Google's Gemini, influence human detection performance.
%Third, since we exclusively recruited participants from one freelance website, our key findings may not be representative of broader populations, such as those found on platforms like Amazon Mechanical Turk or other crowdsourcing services.

\section{Acknowledgments}
This work was in part supported by NSF awards \#1820609,
\#2114824, and \#2131144, and PSU CSRE seed grant 2023. We appreciate our study participants for delivering the results. We also thank the anonymous reviewers for their valuable feedback.

\bibliography{aaai2026}

\section{Ethical Statement}
Our research protocol was approved by the Institutional Review Board (IRB) at our institution. We exclusively recruited participants aged 18 years or older and ensured they were fully informed about the nature of the study. We also explicitly notified them that the dialogue histories and survey responses collected during the study would be shared publicly upon manuscript acceptance. To safeguard privacy, any personally identifiable information, such as names and email addresses, will be removed, and participants will be assigned numerical identifiers to ensure anonymity. In line with the FAIR principles (Findable, Accessible, Interoperable, and Reusable), the data will be managed by Github. 

Before the experiment, we explicitly informed participants that the presented articles, including one of the three paragraphs, contained deepfake content. Consequently, we believe that exposure to these news articles with deepfake paragraphs is unlikely to have negatively influenced the participants. Participants were compensated regardless of whether they completed the entire task and were compensated at rates exceeding the minimum wage.

\section{Paper Checklist}

\begin{enumerate}
    \item  Would answering this research question advance science without violating social contracts, such as violating privacy norms, perpetuating unfair profiling, exacerbating the socio-economic divide, or implying disrespect to societies or cultures?
    \answerYes{Yes}
  \item Do your main claims in the abstract and introduction accurately reflect the paper's contributions and scope?
    \answerYes{Yes}
   \item Do you clarify how the proposed methodological approach is appropriate for the claims made? 
    \answerYes{Yes}
   \item Do you clarify what are possible artifacts in the data used, given population-specific distributions?
    \answerYes{Yes}
  \item Did you describe the limitations of your work?
    \answerYes{Yes}
  \item Did you discuss any potential negative societal impacts of your work?
    \answerYes{Yes}
    \item Did you discuss any potential misuse of your work?
    \answerYes{Yes}
    \item Did you describe steps taken to prevent or mitigate potential negative outcomes of the research, such as data and model documentation, data anonymization, responsible release, access control, and the reproducibility of findings?
    \answerYes{Yes}
  \item Have you read the ethics review guidelines and ensured that your paper conforms to them?
    \answerYes{Yes}

  \item Did you specify all the training details (e.g., data splits, hyperparameters, how they were chosen)? 
    \answerYes{Yes}
     \item Did you report error bars (e.g., with respect to the random seed after running experiments multiple times)?
    \answerNA{Not applicable, we examined generation quality manually. }
	\item Did you include the total amount of compute and the type of resources used (e.g., type of GPUs, internal cluster, or cloud provider)? \answerYes{Yes}
     \item Do you justify how the proposed evaluation is sufficient and appropriate to the claims made? \answerYes{Yes}
     \item Do you discuss what is ``the cost`` of misclassification and fault (in)tolerance?  \answerYes{Yes}

  \item If your work uses existing assets, did you cite the creators? \answerYes{Yes}
  \item Did you mention the license of the assets?
    \answerNo{No, because the license of the assets are mentioned within the cited paper}
  \item Did you include any new assets in the supplemental material or as a URL?
    \answerNo{No, because we mentioned that the dataset and source code will be made publicly available upon acceptance of the manuscript.}
  \item Did you discuss whether and how consent was obtained from people whose data you're using/curating?
    \answerYes{Yes}
  \item Did you discuss whether the data you are using/curating contains personally identifiable information or offensive content?
    \answerYes{Yes}
\item If you are curating or releasing new datasets, did you discuss how you intend to make your datasets FAIR (see \citet{fair})?
\answerYes{Yes}
\item If you are curating or releasing new datasets, did you create a Datasheet for the Dataset (see \citet{gebru2021datasheets})? 
\answerNo{Not yet, as it is a simple and small dataset and will not be publicly shared upon acceptance. However, when we release it publicly, we will create one.}
  \item Did you include the full text of instructions given to participants and screenshots?
    \answerYes{Yes}
  \item Did you describe any potential participant risks, with mentions of Institutional Review Board (IRB) approvals?
    \answerYes{Yes}
  \item Did you include the estimated hourly wage paid to participants and the total amount spent on participant compensation?
    \answerYes{Yes}
   \item Did you discuss how data is stored, shared, and deidentified?
   \answerYes{Yes}
\end{enumerate}

\appendix
\section{Appendix}

\subsection{DeepFakeDeLiBot Training Details}
%\jy{Neo, you can put data stats for finetuning flan-t5 and which compute you used in this section!}

\begin{table}[h!]
\centering
\footnotesize
\begin{tabular}{p{0.09\textwidth}|p{0.08\textwidth}|p{0.09\textwidth}|p{0.12\textwidth}}
\toprule
\textbf{Data}          & \textbf{Dialogues} & \textbf{Avg. Turns} & \textbf{Avg. Probing} \\ 
\midrule
DeliData               & 500                & 28                  & 3.488                 \\ 
Transcribed           & 5                  & 1044.0              & 114.4                 \\ 
Pilot         & 10                 & 224.3               & 37.2                  \\ 
\bottomrule
\end{tabular}
\vspace{1mm}
\caption{Summary of datasets with statistics on dialogues, average turns, and average probing.}
\label{tab:dataset_statistics}
\vspace{-3mm}
\end{table}

For fine-tuning the Flan-T5 Base model, we constructed a synthetic dataset using GPT-3.5-Turbo with in-context learning examples. After manually reviewing and filtering 500 generated data points, we retained 371 for training, 46 for validation, and 47 for testing. Learning rate was set to 5e-5 with minibatch size 2 and 0.01 weight decay using Adam optimizer. The fine-tuning process was conducted on a single Quadro RTX 8000 GPU with 48 GB of memory. The model was trained for 3 epochs, completing within approximately 1 GPU hour, and achieved satisfactory performance.

% \usepackage{booktabs}
\begin{table*}[]
\centering
\footnotesize
\begin{tabular}{@{}c|l@{}}
\toprule
\textbf{Survey Type} & \multicolumn{1}{c}{\textbf{Questions}}                                                                                                                                                                                                                                                                                                                                                                                                                                                                                                                                                                                                                                                                                                                                                                                                                                                                                                                                                                                                                                              \\ \midrule
Pre-Study            & \begin{tabular}[c]{@{}l@{}}- What is your gender?\\ - Which category below includes your age?\\ - Which race/ethnicity best describes you?\\ - What is the highest level of school you have completed?\\ - Please briefly describe your occupation in one or two sentences.\\ - Please rate your self-perceived proficiency in writing on a scale of 1 to 5, \\ with 1 being not proficient at all and 5 being highly proficient.\\ - Have you worked on a similar project before? If not, please insert "N/A". Else, please describe.\\ - Have you ever used AI-powered tools before? If so, how often do you use them?\\ - On a scale of 1 to 5, with 1 being "Not Trusting at All" and 5 being "Highly Trusting", how much would \\ you say you trust AI-powered tools in general?\\ - On a scale of 1 to 5, with 1 being "Very Easy" and 5 being "Very Difficult", please rate the overall difficulty\\  level of this task.\\ - On a scale of 1 to 5, with 1 being "Poorly" and 5 being "Exceptionally Well", how well do you believe you \\ performed this task?\end{tabular} \\ \midrule
Post-Study           & \begin{tabular}[c]{@{}l@{}}- How well do you believe you performed this task?\\ - To what extent did group collaboration benefit your ability to accomplish the task?\\ - In what ways did group collaboration benefit your ability to accomplish the task?\\ - In what ways did group collaboration not benefit your ability to accomplish the task?\\ - During the experiment, did you engage with DEliBot (i.e., a deliberation enhancing dialogue agent)?\\ - How would you rate the quality of Delibot's probing utterances?- How would you describe your overall \\ experience with Delibot's engagement frequency during your interactions?\\ - If you interacted with DEliBot during the experiment, to what extent did the DEliBot benefit the group \\ collaboration?\\ - On a scale of 1 to 5, with 1 being "Not Trusting at All" and 5 being "Highly Trusting", \\ how much would you say you trust DEliBot?\\ - Kindly provide any suggestions you may have for improving DEliBot.\end{tabular}                                                                        \\ \bottomrule
\end{tabular}
\caption{Questions for pre-study and post-study surveys}
\label{tab:survey}
\end{table*}

\begin{table*}[]
\centering
\footnotesize
\begin{tabular}{@{}c|c|c|l@{}}
\toprule
\textbf{Model}            & \textbf{Title}                                                                                                                                            & \textbf{\begin{tabular}[c]{@{}c@{}}Paragraph \\ Number\end{tabular}} & \multicolumn{1}{c}{\textbf{Content}}                                                                                                                                                                                                                                                                                                                                                                \\ \midrule
                          &                                                                                                                                                           & P1                                                                   & \begin{tabular}[c]{@{}l@{}}If Donald Trump was seen as the public face of the failed government \\ response to the coronavirus pandemic, Andrew Cuomo was seen by some \\ as the opposite -- a politician who understood the myriad challenges created \\ by Covid-19 and moved quickly to address them in the most transparent \\ way possible.\end{tabular}                                       \\
                          &                                                                                                                                                           &                                                                      &                                                                                                                                                                                                                                                                                                                                                                                                     \\
                          &                                                                                                                                                           & {\color[HTML]{3531FF} \textbf{P2}}                                   & {\color[HTML]{3531FF} \begin{tabular}[c]{@{}l@{}}One day, Cuomo took the podium at a state event at the hospital \\ where an Ebola patient was being treated.\end{tabular}}                                                                                                                                                                                                                         \\
                          &                                                                                                                                                           &                                                                      &                                                                                                                                                                                                                                                                                                                                                                                                     \\
\multirow{-5}{*}{GPT-2}   & \multirow{-5}{*}{\begin{tabular}[c]{@{}c@{}}Andrew Cuomo's \\ Covid-19 \\ performance may have \\ been less stellar than \\ it seemed\end{tabular}}       & P3                                                                   & \begin{tabular}[c]{@{}l@{}}It was, for many, a refreshing palate cleanser from the obfuscation, spin \\ and denialism that defined how Trump and his administration responded to \\ the virus through the spring and summer of 2020.\end{tabular}                                                                                                                                                   \\ \midrule
                          &                                                                                                                                                           & {\color[HTML]{3531FF} \textbf{P1}}                                   & {\color[HTML]{3531FF} \begin{tabular}[c]{@{}l@{}}House Republicans who voted to impeach former President Donald Trump\\ are facing intense backlash from GOP voters in their home districts, \\ putting their 2022 primaries in jeopardy. \\ The backlash highlights the continued influence of Trump in \\ Republican politics and raises questions about the loyalty of GOP voters.\end{tabular}} \\
                          &                                                                                                                                                           &                                                                      &                                                                                                                                                                                                                                                                                                                                                                                                     \\
                          &                                                                                                                                                           & P2                                                                   & \begin{tabular}[c]{@{}l@{}}The backlash has turned their 2022 primaries into tests of how long \\ Trump can hold the stage in Republican politics and \\ whether GOP voters are willing to turn the midterms into tests of loyalty\\ to him.\end{tabular}                                                                                                                                           \\
                          &                                                                                                                                                           &                                                                      &                                                                                                                                                                                                                                                                                                                                                                                                     \\
\multirow{-5}{*}{GPT-3.5} & \multirow{-5}{*}{\begin{tabular}[c]{@{}c@{}}'People are angry': \\ House Republicans who voted\\  to impeach Trump face \\ backlash at home\end{tabular}} & P3                                                                   & \begin{tabular}[c]{@{}l@{}}The group of 10 Republicans includes moderates in swing districts, as well as \\ some reliable conservatives, including the No. 3-ranking House Republican, \\ Wyoming Rep. Liz Cheney, and South Carolina Rep. Tom Rice\end{tabular}                                                                                                                                    \\ \bottomrule
\end{tabular}
\caption{Deepfake article example. Text in \textcolor[HTML]{3531FF}{blue} indicates a paragraph written by LLMs.}
\label{tab:deepfake-example}
\end{table*}

\begin{table*}[]
    \centering
    \footnotesize
\begin{tabular}{@{}c|c|c|c|c|c|c@{}}
\toprule
\textbf{Main Features} & \textbf{Coef} & \textit{\textbf{p}}          &  & \textbf{Interaction Features} & \textbf{Coef} & \textit{\textbf{p}}                            \\ \cmidrule(r){1-3} \cmidrule(l){5-7} 
Q7                     & 0.66               & 0.89                       &  & Q7\_DeepFakeDeLiBot  & 3.88      & 0.54                       \\
Q9                     & 4.37             & 0.21                        &  & Q9\_DeepFakeDeLiBot  & 2.53      & 0.54                        \\
Q10                    & 7.35               & 0.14 &  & Q10\_DeepFakeDeLiBot & 0.54       & 0.96                       \\
Q2                     & 0.26              & 0.96                      &  & Q2\_DeepFakeDeLiBot  & 2.02      & 0.81                       \\
Q3                     & 3.34               & 0.27                        &  & Q3\_DeepFakeDeLiBot  & 9.64     & {\color[HTML]{FE0000} 0.03} \\ \bottomrule
\end{tabular}
\vspace{1mm}
\caption{Linear regression results of participants' background/experiences and detection performance gain. Investigated features include: self-perceived proficiency in writing (Q7), AI-powered tool usage levels (Q9), and their trust levels in AI-powered tools (Q10),their self-perceived performance after group collaboration (Q2), and self-perceived effectiveness of group collaboration (Q3).}
\label{tab:rq3_results}
\vspace{-1mm}
\end{table*}

\begin{table*}[]
\centering
\footnotesize
\begin{tabular}{@{}c|c|c|c|c|c|c@{}}
\toprule
\textbf{Main Features}                                & \textbf{Coef} & \textit{\textbf{p}}          &                       & \textbf{Interaction Features}                                                                         & \textbf{Coef} & \textit{\textbf{p}}                            \\ \cmidrule(r){1-3} \cmidrule(l){5-7} 
Participant engagement                                & -0.007              & 0.47                         &                       & \begin{tabular}[c]{@{}c@{}}Participant engagement\\ \_DeepFakeDeLiBot\end{tabular}           & 0.005      & 0.67                        \\
Even participation                                    & -0.26              & 0.56                       &                       & \begin{tabular}[c]{@{}c@{}}Even participation\\ \_DeepFakeDeLiBot\end{tabular}               & -0.39     & 0.59                       \\
Consensus formation                                   & 0.006               & {\color[HTML]{000000} 0.89} &                       & \begin{tabular}[c]{@{}c@{}}Consensus formation\\ \_DeepFakeDeLiBot\end{tabular}              & 0.05      & 0.55                        \\
Solution probing frequency                            & 0.52               & 0.42                   &                       & \begin{tabular}[c]{@{}c@{}}Solution probing frequency\\ \_DeepFakeDeLiBot\end{tabular}       & 0.60      & 0.62                        \\
Reasoning probing frequency                           & -0.59              & 0.44                        &                       & \begin{tabular}[c]{@{}c@{}}Reasoning probing frequency\\ \_DeepFakeDeLiBot\end{tabular}      & 0.63      & {\color[HTML]{000000} 0.5} \\
\multicolumn{1}{l|}{Moderation probing frequency}     & 0.48               & 0.14                        & \multicolumn{1}{l|}{} & \begin{tabular}[c]{@{}c@{}}Moderation probing frequency\\ \_DeepFakeDeLiBot\end{tabular}     & -0.9     & {\color[HTML]{FE0000} 0.05}  \\
\multicolumn{1}{l|}{Diversity of discussed solutions} & 0.004               & 0.27                        & \multicolumn{1}{l|}{} & \begin{tabular}[c]{@{}c@{}}Diversity of discussed solutions\\ \_DeepFakeDeLiBot\end{tabular} & -0.02     & 0.72                        \\
\multicolumn{1}{l|}{Diversity of submitted reasoning} & 0.006               & 0.76                        & \multicolumn{1}{l|}{} & \begin{tabular}[c]{@{}c@{}}Diversity of submitted reasoning\\ \_DeepFakeDeLiBot\end{tabular} & -0.0007     & 0.98                        \\ \bottomrule
\end{tabular}
\vspace{1mm}
\caption{Linear regression results of group dynamics and detection performance gain.}
\label{tab:rq3_result2}
\vspace{-3mm}
\end{table*}

% \usepackage{booktabs}
% \usepackage[normalem]{ulem}

\begin{table*}[]
\centering
\footnotesize
\begin{tabular}{@{}lll@{}}
\toprule
\multicolumn{1}{c}{\textbf{Code Category}} & \multicolumn{1}{c}{\textbf{Description}}                                                                                                      & \multicolumn{1}{c}{\textbf{Example Bot Utterance}}                                                                                                                                                                                                                                                          \\ \midrule
\textbf{Temporal Misalignment}             & \begin{tabular}[c]{@{}l@{}}Bot prompt is mistimed—occurs after group \\ consensus or  before relevant context is \\ established.\end{tabular} & {\color[HTML]{333333} \begin{tabular}[c]{@{}l@{}}Zebra: So we’re all good with Paragraph 2?\\ Penguin: Yep.\\ Bot: What is your reason for choosing Paragraph 3?\\ Zebra: We already picked 2...\\ Penguin: Yeah, too late now.\end{tabular}}                                                            \\ \midrule
\textbf{Dismissal}                         & \begin{tabular}[c]{@{}l@{}}Participants ignore or reject the prompt \\ without elaboration.\end{tabular}                                      & \begin{tabular}[c]{@{}l@{}}Bot: What do you think about Paragraph 1?\\ Penguin: Anyway, 3 is the weirdest one. Let's move on.\\ Zebra: Sure.\end{tabular}                                                                                                                                                \\ \midrule
\textbf{Task Acceleration}                 & \begin{tabular}[c]{@{}l@{}}Group moves too quickly for bot to intervene \\ effectively.\end{tabular}                                          & \begin{tabular}[c]{@{}l@{}}Zebra: Let’s just pick Paragraph 1. It’s probably fake.\\ Penguin: Agreed. Done.\\ Zebra: What about for the next question? I chose 3.\\ Penguin: Same, let's move on.\end{tabular}                                                                                              \\ \midrule
\textbf{Contextual Receptiveness}          & \begin{tabular}[c]{@{}l@{}}Group engages with the prompt, elaborates or \\ builds on the idea  in meaningful ways.\end{tabular}               & \begin{tabular}[c]{@{}l@{}}Zebra: Paragraph 2 feels kind of off, but I can’t tell why.\\ Bot: What makes Paragraph 2 stand out to you?\\ Penguin: It’s more narrative—it reads like a story.\\ Zebra: Yeah. Also, that paragraph talks about Ebola while\\  the other two talk about Covid.\end{tabular} \\ \midrule
\textbf{Prompt Relevance}                  & \begin{tabular}[c]{@{}l@{}}Bot's utterance matches the group’s current \\ topic or reasoning  trajectory.\end{tabular}                        & \begin{tabular}[c]{@{}l@{}}Penguin: I chose Paragraph 1 as the answer due to some \\ logical gaps.\\ Zebra: I chose Paragraph 1 because of grammatical errors.\\ Bot: Do we all agree that Paragraph 1 is the answer?\\ Penguin: Yes!\\ Zebra: Yup.\end{tabular}                                         \\ \midrule
\textbf{Prompt Quality}                    & \begin{tabular}[c]{@{}l@{}}Prompt is clear, focused, and designed to \\ elicit deeper reasoning.\end{tabular}                                 & \begin{tabular}[c]{@{}l@{}}Penguin: I think Paragraph 1 is trying too hard to be \\ emotional.\\ Bot: Can you further explain on that?\\ Penguin: Yeah, it’s full of dramatic language. That’s not\\  how news sounds.\\ Zebra: Good catch. Real articles are usually more neutral.\end{tabular}         \\ \bottomrule
\end{tabular}
\caption{Dialogue annotation scheme and examples.} %\jy{will double-check the numbers and update the table}
\label{tab:code}
\end{table*}

\end{document}